\documentclass[acmsmall,screen]{acmart}
\renewcommand\footnotetextcopyrightpermission[1]{} 
\AtBeginDocument{%
  \providecommand\BibTeX{{%
    \normalfont B\kern-0.5em{\scshape i\kern-0.25em b}\kern-0.8em\TeX}}}

\setcopyright{acmcopyright}
\copyrightyear{2022}
\acmYear{2022}
\acmDOI{10.1145/1122445.1122456}

\acmJournal{JACM}
\acmVolume{37}
\acmNumber{4}
\acmArticle{111}
\acmMonth{8}

\usepackage{indentfirst}
\usepackage{color}
\usepackage{url}
\usepackage[T1]{fontenc}
\usepackage{lmodern}

\begin{document}

\title{Aesthetic Attribute Assessment of Images Numerically on Mixed Multi-attribute Datasets}

\author{Xin Jin}
\email{jinxinbesti@foxmail.com}
\affiliation{%
	\institution{Beijing Electronic Science and Technology Institute}
	\city{Beijing}
	\state{Beijing}
	\country{China}
}

\author{Xinning Li}
\email{l_xinning@163.com}
\affiliation{%
	\institution{Beijing Electronic Science and Technology Institute}
	\city{Beijing}
	\state{Beijing}
	\country{China}
}

\author{Hao Lou$^{*}$}
\email{452392771@qq.com}

\affiliation{%
	\institution{Beijing Electronic Science and Technology Institute}
	\city{Beijing}
	\state{Beijing}
	\country{China}
}

\author{Chenyu Fan}
\email{3497961491@qq.com}
\affiliation{%
	\institution{Beijing Electronic Science and Technology Institute}
	\city{Beijing}
	\state{Beijing}
	\country{China}
}

\author{Qiang Deng}
\email{1352110584@qq.com}
\affiliation{%
	\institution{Beijing Electronic Science and Technology Institute}
	\city{Beijing}
	\state{Beijing}
	\country{China}
}

\author{Chaoen Xiao$^{*}$}
\email{xcecd@qq.com}
\thanks{*Corresponding authors}
\affiliation{%
	\institution{Beijing Electronic Science and Technology Institute}
	\city{Beijing}
	\state{Beijing}
	\country{China}
}

\author{Shuai Cui}
\email{shucui@ucdavis.edu}
\affiliation{%
	\institution{University of California, Davis}
	\city{Davis}
	\country{USA}
}

\author{Amit Kumar Singh}
\email{amit.singh@nitp.ac.in}
\affiliation{%
	\institution{National Institute of Technology Patna}
	\country{India}
}



\begin{abstract}
With the continuous development of social software and multimedia technology, images have become a kind of important carrier for spreading information and socializing. How to evaluate an image comprehensively has become the focus of recent researches. The traditional image aesthetic assessment methods often adopt single numerical overall assessment scores, which has certain subjectivity and can no longer meet the higher aesthetic requirements. In this paper, we construct an new image attribute dataset called aesthetic mixed dataset with attributes(AMD-A) and design external attribute features for fusion. Besides, we propose a efficient method for image aesthetic attribute assessment on mixed multi-attribute dataset and construct a multitasking network architecture by using the EfficientNet-B0 as the backbone network. Our model can achieve aesthetic classification, overall scoring and attribute scoring. In each sub-network, we improve the feature extraction through ECA channel attention module. As for the final overall scoring, we adopt the idea of the teacher-student network and use the classification sub-network to guide the aesthetic overall fine-grain regression. Experimental results, using the MindSpore, show that our proposed method can effectively improve the performance of the aesthetic overall and attribute assessment.
\end{abstract}

\ccsdesc[500]{Applied computing~Media arts}

\keywords{aesthetic mixed dataset with attributes, multitasking, external attribute features, ECA channel attention}

\begin{teaserfigure}
	\includegraphics[width=\textwidth]{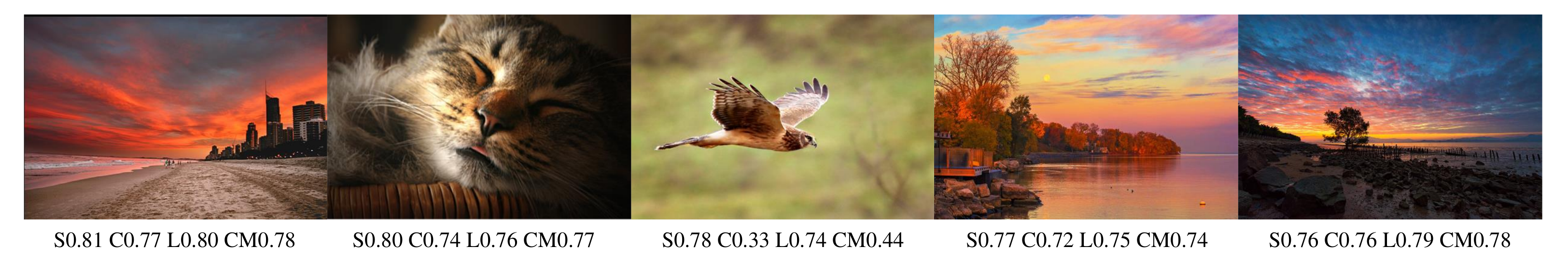}
	\caption{The examples for image aesthetic attribute assessment, S represents the overall score, C represents the color score, L represents the light score, and CM represents the composition score}
	\label{fig:fig1}
\end{teaserfigure}
\maketitle

\section{Introduction}
The visual aesthetic quality of the image measures the visual attraction of the images for humans. Since visual aesthetics is a subjective attribute \cite{kairanbay2019beauty}, it always depends on personal emotions and preferences. This makes image aesthetic quality assessment a subjective task and evaluated only by experts. If there is a large number of image samples, the efficiency of artificial aesthetic quality assessment will be quite low. However, people tend to agree that some images do indeed more attractive than others in daily life, which creates a computational aesthetics \cite{2008STUDYING}. Computational aesthetics lets the computer mimic the process of human aesthetic assessment and compute the methods to predict the aesthetic quality of the images automatically. 

The focus of computational aesthetics is to predict people's emotional responses to visual stimuli through computational technology, study the internal mechanism of human perception and explore the mystery of artificial aesthetic intelligence. Computational visual aesthetics \cite{2017Computational} is the computational processing of human visual information. Image aesthetic assessment is the most popular research direction in the field of computational visual aesthetics, and it is also the first step in studying computational visual aesthetics. Image aesthetic assessment is to simulate human perception and cognition by the computers, which can provide the quantitative assessment of aesthetic quality. Image aesthetic attributes assessment mainly focuses on the quantitative assessment formed by aesthetic attributes such as composition, color and light in the photographic images.

There are two main parts for the image aesthetics quality assessment: the feature extraction part and the assessment part. In the feature extraction, Yan et al. \cite{1640788} proposed the 7-dimensional aesthetic features based on photography knowledge and high-level semantics. The features include simplicity, spatial edges distribution, color distribution, hue count, blur, contrast and brightness. With the development of deep learning, researchers introduce deep convolutional neural networks in the task of image aesthetics assessment. Due to the ability of learning features automatically, people can extract aesthetic features from images by the deep convolutional neural networks without a lot of aesthetic knowledge and professionally photography experiences. However, there are deficiencies in general aesthetic benchmark datasets. For example, in AADB \cite{2016Photo}, each image is only marked by a small number of annotation. So the data labels is kind of subjective. It is difficult to extract the aesthetic attribute features and limits the extraction ability of features for the model. At the same time, the multidimensional attribute assessment will lead to a sharp increase in the number of network parameters, which is not conducive to the actual development and application.

To solve these problems, we propose a efficient method for image aesthetic attribute assessment based on the mixed multi-attribute dataset. Our work includes the following three contributions:

1. We screen and construct the aesthetic mixed dataset with attributes(AMD-A), which has more image aesthetic attribute annotations compared with the traditional aesthetic attribute datasets.

2. We propose an aesthetic multitasking network architecture based on EfficientNet-B0 and ECA channel attention modules to simplify the model parameters and realize the aesthetic classification, overall scoring and attribute scoring.  

3. We design several kinds of external attribute features and use feature fusion to imporve the performance of the image aesthetic attribute assessment. Besides, inspired by teacher-student network, we propose the soft loss to imporve the performance of the image aesthetic overall assessment.  

\section{Related work}

\subsection{Image aesthetic assessment}
Image aesthetic assessment is usually regarded as a classification or score regression task. For the classification task, images can generally be divided into high-quality images and low-quality images \cite{1969Interhemispheric}. For the regression task, images can be evaluated according to aesthetic overall scores. However, the overall scores can not accurately measure the results of aesthetic assessment, which largely ignores the diversity, subjectivity, and personality in the human aesthetic consensus. An image also contains many aesthetic attributes such as light and shadow, color, composition, blur, movement, and interest. So the related work has focused on the multidimensional aesthetic attribute assessment.

In the early stage, the handcraft designing features was the main way to extract features from images \cite{2004Classification,2008Photo,2011Aesthetic,2011Assessing}. Datta et al. \cite{2008STUDYING} used bottom-level features (color, texture, shape, image size, etc.) and high-level features (depth of field, rule of thirds, regional contrast) as the image aesthetic features. Marchesotti \cite{6126444} and others directly used SIFT (BOV or FisherVector) and local color descriptors to classify aesthetic images. 

Today, deep learning has been widely used in different walks of life, such as IoV \cite{wang2022software}, IoT \cite{wang2022time} and edge intelligence \cite{zhang2020psac}. With the development of deep learning, the researchers introduce deep convolutional neural networks into the image aesthetic assessment. Due to the powerful learning capabilities in deep convolutional neural networks, people can automatically extract aesthetic features without substantial theoretical knowledge and photographic experience. In recent years, deep convolutional neural networks have shown good performance in the overall regression and classification tasks of image aesthetics, and a series of excellent models \cite{2017A,2015Visual,2017Aestheticd,2021Hierarchical,zhou2022double,xu2022mfgan,zhen2022towards} have emerged. At the same time, in the field of image aesthetic attribute assessment, there are image aesthetic attribute scoring methods based on hierarchical multitasking networks \cite{2018Predicting} and incremental learning multitasking networks \cite{2019Incremental} by using image composition, light and color, exposure depth of field and other attributes. However, due to the limitation of data quantity and the categories of aesthetic attributes, the above attribute assessment method don't have high accuracy. With the increasing of attribute categories, the network models become extremely complex, which is not conducive to expansion and practical application.

\subsection{Aesthetic attribute datasets}
The emergence of large-scale aesthetic datasets provides a rich source of samples for aesthetic assessment model training. Murray et al. \cite{2012AVA} constructed an aesthetic visual analysis dataset(AVA), containing 255,530 images from the www.dpchallenge.com. AVA is the benchmark dataset for the current image aesthetic quality assessment task. Each image includes a numerical aesthetic overall regression label, 66 semantic labels, and 14 style labels. Building on this dataset, Kang et al. \cite{kang2020eva} proposed the explainable visual aesthetics dataset(EVA), which contains 4070 images. Each image has 30 annotation scores at least and mainly includes an overall score and 4 different aesthetic attribute scores. Compared with AVA, EVA adopts a more rigorous annotation collection method and overcomes the noisy labels due to the misunderstanding bias of the annotators, which facilitates the research on aesthetic understanding.

Kong et al. \cite{2016Photo} designed the aesthetics and attributes database(AADB). The dataset contains 9958 images from professional photographers and ordinary photographers, each image has an overall score and eleven aesthetic attributes scores. Chang et al. \cite{2017Aesthetic} first proposed the annotation information for aesthetic language assessment, and designed the photo critique captioning dataset(PCCD), which contains 4235 valid images from the foreign photography website Gurushots.com. Except for the overall score, each image has language comments and scores on the composition perspective, color illumination, image theme, depth of field, focus, and camera use. Fang et al. \cite{2020Perceptual} conducted the first systematic study on the smartphone image aesthetic quality assessment and constructed smartphone photography attribute and quality dataset(SPAQ). The dataset consists of 11125 images taken by 66 smartphones, and each image including an overall label, an image attribute label and a scene category label.

\subsection{Feature fusion}
Feature fusion is a common way to improve the model performance. In the field of aesthetic assessment, the low-level features obtained by deep neural networks have high resolution and contain more aesthetic low-level information, such as color, texture and structure. But after fewer convolutional layers, low-level features contain lower high-level semantics and are easy to cause noise interference \cite{zhou2022contextual,zhou2019multi}. High-level features tend to have stronger semantic information, but they have very low feature resolution and poor perception of details. In the aesthetic field, local aesthetic attributes based on low-level information are as important as global aesthetic features based on high-level information. Especially in the regression task of aesthetic attributes. The output results of deep convolutional networks are often in the last layer of the network, which leads to the inability to effectively extract the corresponding attribute features in the assessment of aesthetic attributes.

In related work, the researchers improve the performance of detecting and dividing objects by integrating the network layer features at different locations of the neural network. We classified feature fusion and predicted output as early and late fusion. The idea of early fusion strategy is to integrate the high-level and low-leve features, and then conduct model training and prediction on the fusion features. Such methods usually use concatting and adding operations to fuse the features. Related research work such as Inside-Outside Net \cite{2016Inside} and HyperNet \cite{2016HyperNet}. The idea of the late fusion strategy is to output the prediction results of different network layers and then to fuse all the detection results. Related research work such as Feature Pyramid Network \cite{2017Feature}, Single Shot MultiBox Detector \cite{2016SSD}, and Densenet \cite{2016Densely}. This paper mainly adopts the early fusion strategy to integrate different high-level features of the sub-networks and external attribute features to improve the supervised assessment performance of image aesthetic attributes.
    
\section{Aesthetic mixed dataset with attributes(AMD-A)}
In order to construct a dataset with a reasonable distribution both in image aesthetics quality assessment and image aesthetic attributes assessment, we rebuild a dataset named Aesthetic mixed dataset with attributes(AMD-A) cincluding 16924 images. According to different tasks, AMD-A is divided into two sets. One set(11166 images) is applied to aesthetic overall score regression, another set(16924 images) is applied to aesthetic attribute score classification and regression.

As for the attribute regression, we collect images from EVA \cite{kang2020eva}, AADB \cite{2016Photo}, PCCD \cite{2017Aesthetic}, PADB, and HADB. PADB and HADB are self-built datasets. Each image has three attribute labels and one overall label for assessment. The three attributes include light, color and composition. To increase the training samples of the aesthetic regression, we collected 5758 images with only overall scores from other datasets. There are  from DPCallenge.com, SCUI-FBP5500 \cite{2018SCUT}, Photo.net and SPAQ \cite{2020Perceptual}. Data labels are continuous and each one has a score range of 0-1. The distribution of different labels are as Fig.2. 

\begin{figure}[ht]
	\centering
	\includegraphics[width=\textwidth]{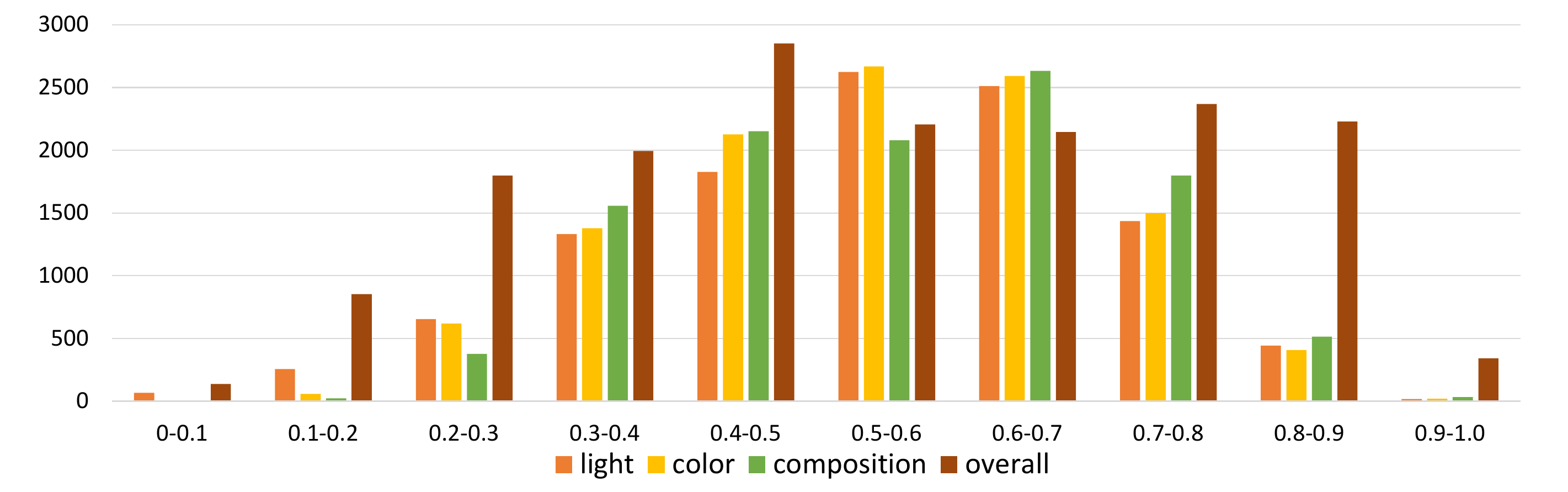}
	\caption{The distribution of light, color, composition and overall labels in AMD-A}
\end{figure}

We calculated the means and the standard deviations of the labels in Table 1. The results show that the 4 different categories of labels have the similar data distributions.

\begin{table}[htbp]
	\caption{Means and standard deviations for light, color, composition and overall labels}
	\centering
	\label{tab:1}
	\begin{tabular}{ccccc}
		\hline\noalign{\smallskip}
		 & light  & color & composition & overall \\
		\noalign{\smallskip}\hline\noalign{\smallskip}
		Mean & 0.5324 & 0.5472 & 0.5420 & 0.5398   \\
		Standard deviation & 0.1522 & 0.1319 & 0.1388 & 0.2116   \\
		\noalign{\smallskip}\hline
	\end{tabular}
\end{table}

\section{Network architecture}
We build a multitasking network architecture including the backbone network and five sub-networks. The basic network architecture is shown in Fig.3.

\begin{figure}[ht]
	\centering
	\includegraphics[width=\columnwidth]{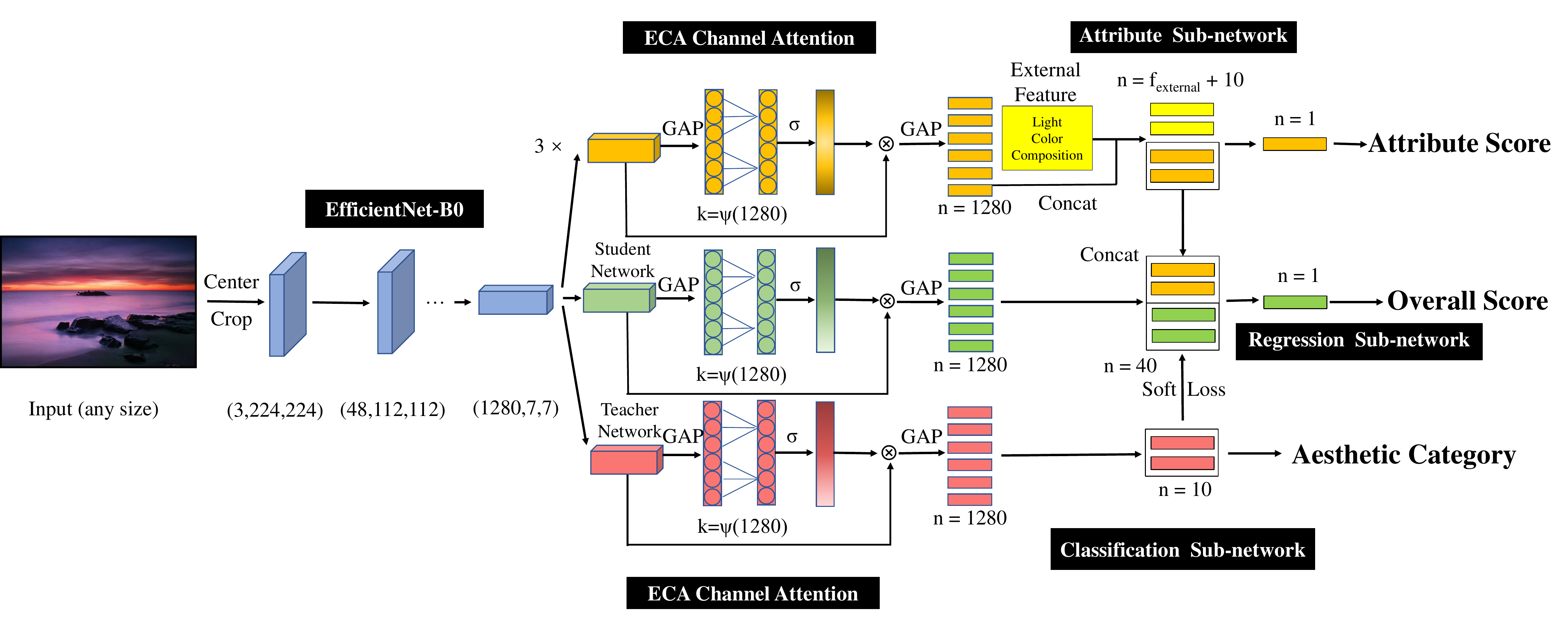}
	\caption{The architecture of the network}
\end{figure}

We use EfficientNet-B0 as the backbone network \cite{2019EfficientNet}. EfficientNet has an efficient feature extraction ability and can achieve more accurate aesthetic regression with a small number of parameters. The size of the parameter model for EffiecinetNet-B0  is only 20M, which is smaller than many existing model parameter model. So it is an ideal pre-trained parameter model in development and application scenarios. We use the blue part for the backbone network in the fig.3.

Between the backbone network and the sub-networks, there are ECA channel attention modules. In ECA, GAP is the global average pooling, $\sigma$ is the sigmoid function, $\psi(C)$ is the function of parameter \textit{k}, and each layer will be activated by the relu function. We will introduce it in Section 4.1.

The five sub-networks are one classification sub-network, three attribute sub-networks and one regression sub-network consisting of fully connected layers. We use the red part for the classification sub-network, the khaki part for the attribute sub-networks and the green part for the regression sub-network. The attribute sub-networks in the figure have the similar structure, so we omitted additional attribute sub-networks and displayed only one attribute sub-network. The yellow part is for the external attribute features(4 light features, 7 color features, and 10 composition features). 

From the architecture of the network, we can see two different kinds of feature fusions. One for external features and 10 attribute features in the attribute regression tasks, and another one for 10 regression features and 30 attribute features in the regression task. The training details will be explained in Section 4.2.

\subsection{ECA channel attention}
From the network structure in Fig.3, each ECA channel attention module is added between the backbone network and each sub-network. Recently, there are a lot of researches improving channel and spatial attention to make progresses in the performance. The ECA channel attention can improve the ability of extracting features. It mainly improves from the SENet \cite{2019ECA} module. ECA channel attention proposes a local cross-channel interaction strategy without the dimension reduction and implement a method of adaptively selecting the size of the one-dimensional convolution kernel. So we use it to improve the performance of the image aesthetic attribute and overall assessment. To the best of our knowledge, it is also the first time to use the ECA channel attention in the task of aesthetics attribute assessment.

Instead of reducing global average pooling layers at the channel level, ECA channel attention captures local cross-channel interaction information by considering each channel and its \textit{k} neighbors. This can be effectively achieved efficiently by a fast one-dimensional convolution. \textit{k} indicates how many neighbors near the channel participate in the attention prediction of this channel. To avoid artificially raising \textit{k} by cross-validation, \cite{2019ECA} proposes a way to determine \textit{k} automatically. The size of the convolutional kernel \textit{k} is proportional to the channel dimension as follows:

\begin{equation}
	k=\psi(C)\ =\left|\frac{\log_2(C)}{\gamma}+\frac{b}{\gamma}\right|_{\mathrm{odd}}	
\end{equation}

${|t|}_{odd}$ means the nearest odd number of \emph{t}, the value of \emph{$\gamma$} is 2, the value of \emph{b} is 1. We can get \emph{k} = 7 if \emph{C} = 1280. 

\subsection{Multitasking network}

We use EfficientNet-B0 as the backbone network and design the classification sub-network for the classification task. Between these two networks, we use the ECA channel attention module to improve the capability of the aesthetic feature extraction. The classification task divides aesthetic images into ten categories based on the level of aesthetic scores, which is a coarse-grained aesthetic regression. During training, we relaxed the parameters of the backbone network and the classification sub-network, obtained the high-level aesthetic features of 1280×7×7 through the EfficientNet-B0, and made the feature reweighted through the ECA channel attention module. After that, we used the global average pooling layer to transform the high-level features into 1280 features, then reduced the features to 10 through one full connection layer and the activation function. Each feature of the output is the probability of each aesthetic category.

\begin{figure}[htbp]
	\centering
	\includegraphics[scale=0.35]{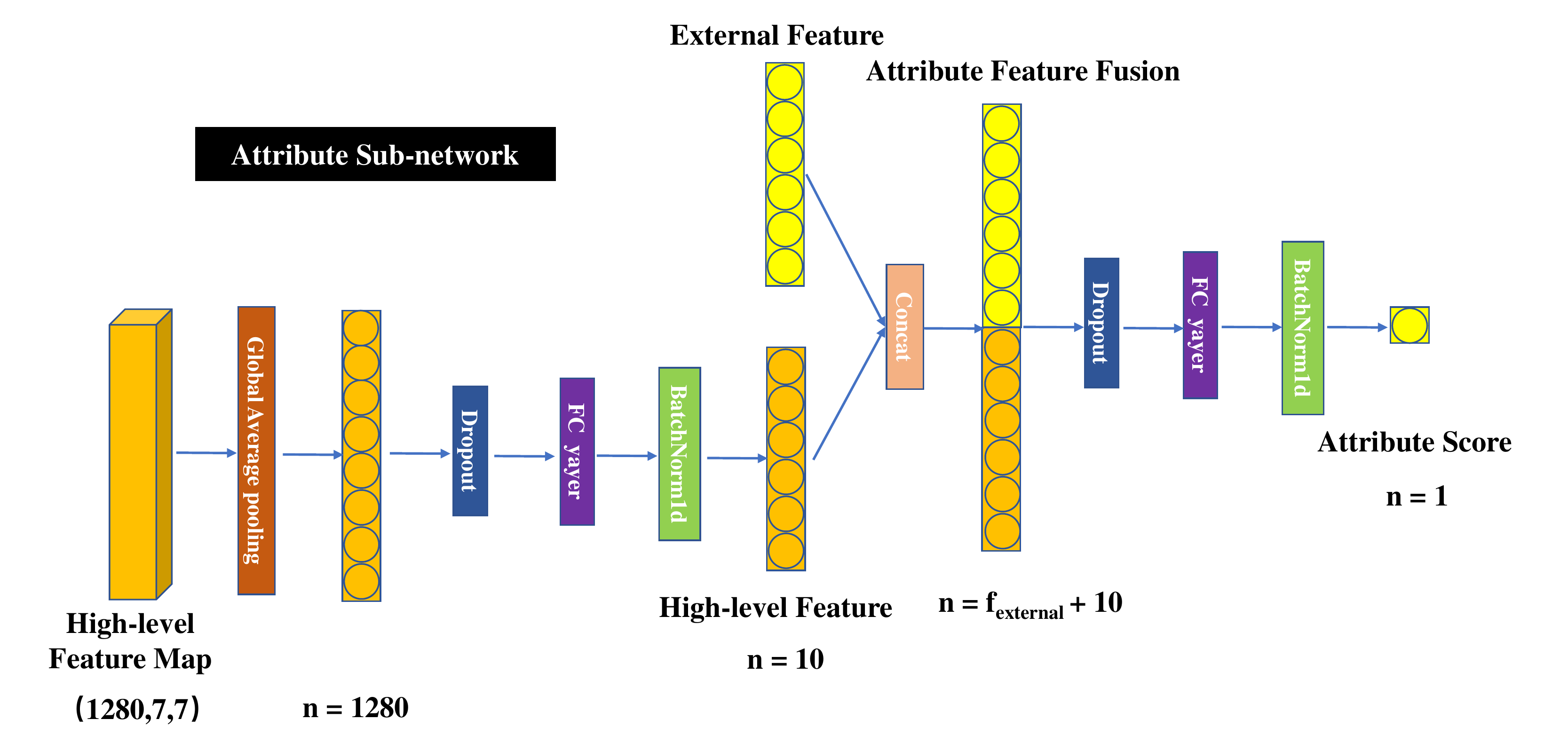}
	\caption{Attribute feature fusion}
\end{figure}

After completing the classification task, we trained the three sub-networks in the order of light, color and composition. The training labels are the attribute labels of each image. The structure of attribute sub-networks are similar to the classification sub-networks. We reduced 1280 features to 10 features by one fully connected layer and the activation function. These 10 features are the attribute features of the image. As shown in Fig.4, we spliced different external features onto attribute features and regress the features to an attribute score using one fully connected layer and the activation function. The design and calculation of the external features will be mentioned in Section 5.

\begin{figure}[htbp]
	\centering
	\includegraphics[scale=0.35]{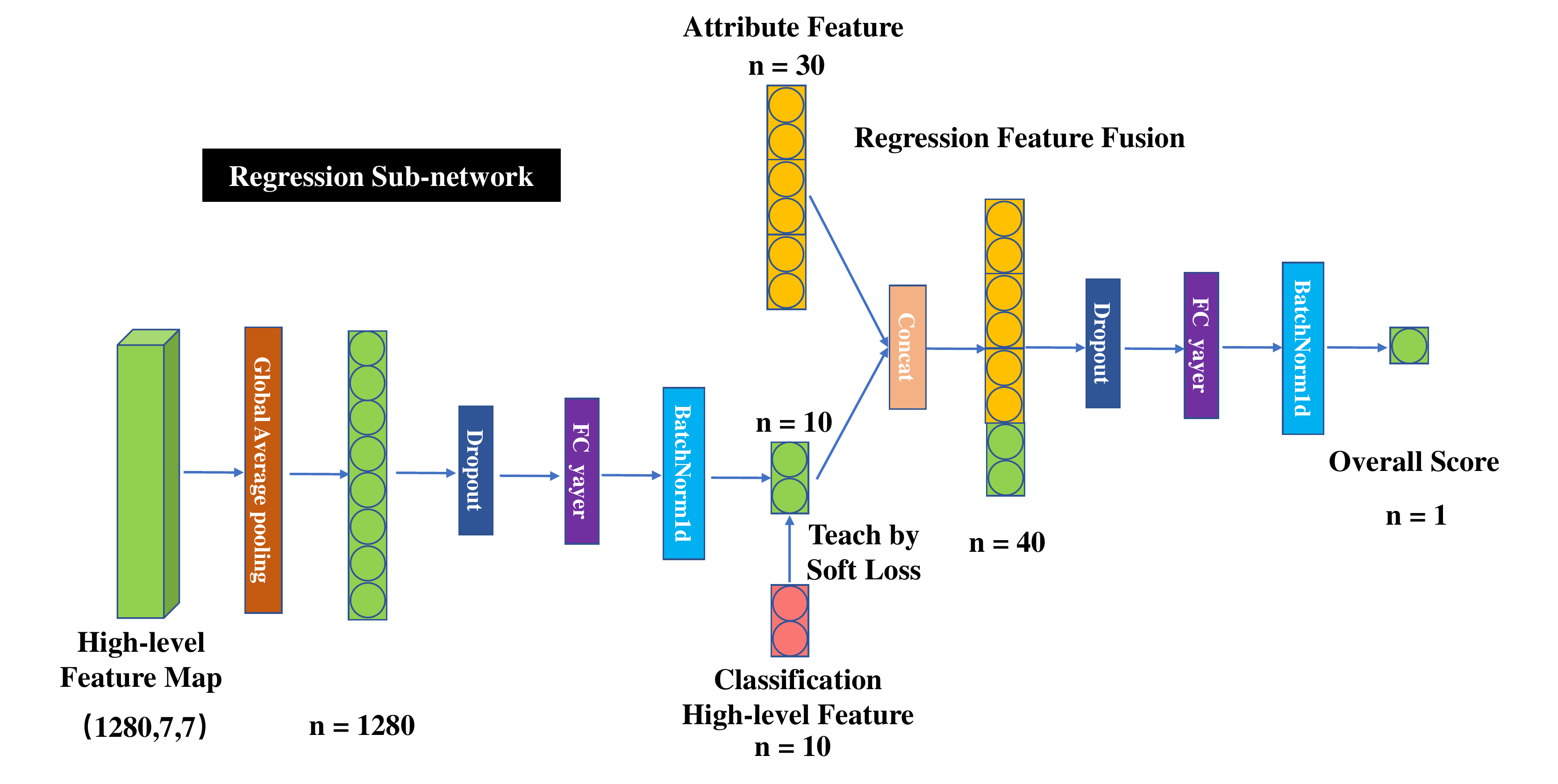}
	\caption{Overall feature fusion}
\end{figure}
\begin{equation}
 Loss = MSE Loss + \lambda × Soft Loss	
\end{equation}

As shown in Fig.5, we also reduced 1280 features to 10 features in the regression sub-network. Then, we combined these 10 features with 3×10 attribute features in the penultimate layer of the attribute sub-network to obtain an aesthetic overall score through one fully connected layer and the activation function. Inspired by the idea of teacher-student networks \cite{2015Distilling}, we used 10 features from the output of the classification sub-network to guide the 10 features of the regression sub-network. We calculate the relative entropy between the two feature sequences and define it as the soft loss. The regression sub-network is gradient-adjusted by the mse loss of the aesthetic score regression and a weighted soft loss. We set the weight of the soft loss as 0.1, which we will prove in Section 6.

\section{Attribute Features Fusion}

Before training the three attribute regression sub-networks, we will design and calculate the external attribute features of the input images. The external attribute features includes light features, color features and composition features. We will store the external attribute features in the data labels. The design details are presented in this section.

\subsection{Light features}
Referring to the value and lightness information \cite{2008STUDYING}, we extract the light features as mean of value(\textit{f1}), standard deviation of value(\textit{f2}), mean of lightness(\textit{f3}) and standard deviation of lightness(\textit{f4}).

The light features was mainly obtained by calculating the mean and standard deviation of the L and V channels. Both the value and lightness range is 0-255. We use x to represent the pixels in the images. The calculation formulas of average brightness (\textit{f1}), standard deviation of brightness (\textit{f2}), average lightness (\textit{f3}) and standard deviation of lightness (\textit{f4}) are as follows:
\begin{equation}
	f_1=\frac{1}{|I|}{\mathrm{\sum}_{x\in{I}}L(x)}	
\end{equation}

\begin{equation}
	f_2=std(L(x))	
\end{equation}

\begin{equation}
	f_3=\frac{1}{|I|}{\mathrm{\sum}_{x\in{I}}V(x)}	
\end{equation}

\begin{equation}
	f_4=std(V(x))	
\end{equation}

\begin{figure}[ht]
	\centering
	\includegraphics[width=\textwidth]{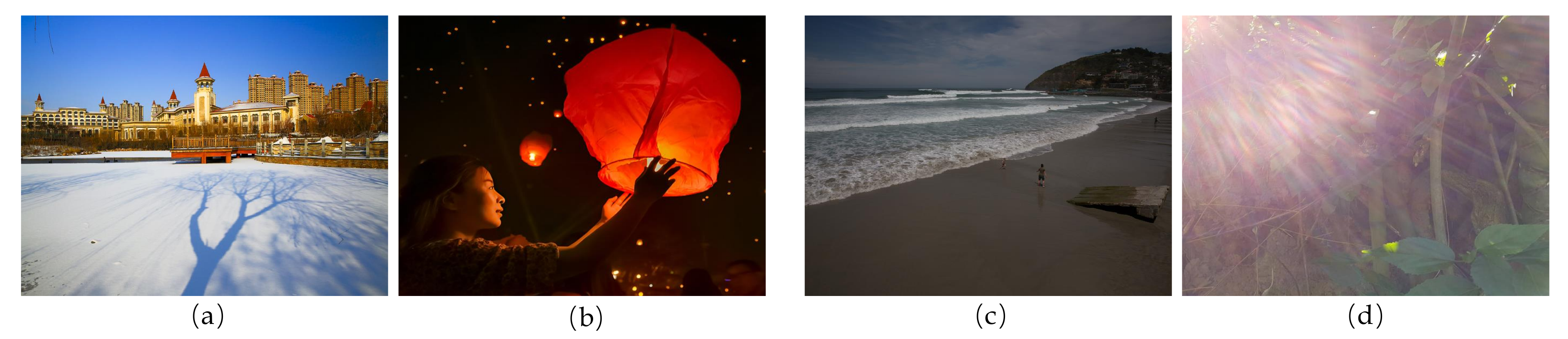}
	\caption{Samples with different light effects}
\end{figure}

\begin{table}[htbp]
	\caption{Light feature values for samples}
	\centering
	\label{tab:1}
	\begin{tabular}{ccccc}
		\hline\noalign{\smallskip}
		Light feature & 6(a)  & 6(b) & 6(c) & 6(d) \\
		\noalign{\smallskip}\hline\noalign{\smallskip}
		Mean of value (\textit{f1}) & 148.253016 & 34.985529 & 72.411960 & 139.248896   \\
		Standard deviation of value (\textit{f2}) & 52.104682 & 40.385166 & 29.037064 & 39.939841   \\
		Mean of lightness (\textit{f3}) & 183.823056 & 148.253016 & 79.659121 & 151.227269  \\
		Standard deviation of lightness (\textit{f4}) & 50.643524 & 78.682690 & 28.775265 & 43.909947   \\
		\noalign{\smallskip}\hline
	\end{tabular}
\end{table}
Fig.6 shows four sample images of different light effects, where Fig.6(a) and Fig.6(b) have better light effects, while Fig.6(c) and Fig.6(d) have poor light effect. Table 2 shows the light feature values corresponding to these four images. From the feature values in Table 2, we can see that the images with better light effect generally have higher means and standard deviations. While the image is too dark, the means and standard deviations are lower, and if the exposure is too excessive, the means will be higher and the standard deviations will be lower.

\subsection{Color features}
Referring to the color information \cite{2012The}, we extract the color features as the weight of color channel (\textit{f1}), the number of RGB dominant colors (\textit{f2}), the degree of RGB dominant colors (\textit{f3}), the number of HSV dominant colors (\textit{f4}), the degree of HSV dominant colors (\textit{f5}), the number of dominant hues (\textit{f6}), the contrast ratio of dominant hues (\textit{f7}).

By calculating the  approximation degree of the RGB color channel, the images can be divided into colored three-channel images, approximate grayscale channel images and grayscale single-channel images. The weight of color channel \textit{f1} is calculated by converting the image into the RGB channels. If the values of the three channel are exactly the same, \textit{f1} = 0. If the three channels are not exactly consistent, we will calculate the average difference between the RGB channels at each pixel, while the average difference is less than 10, \textit{f1} = 0.5; otherwise, the images are considered as the colored three-channel images, and \textit{f1} = 1.

The dominant colors are mainly calculated by the color histogram in the RGB channel and the hue histogram in the HSV channel. As for RGB dominant colors, we quantize each RGB channel to 8 values, so that a 512-dimensional histogram {h$_{RGB}$}=\{{h$_{0}$}, {h$_{1}$},..., {h$_{511}$}\} can be created. {h$_{i}$} represents the number of pixels in the i-th histogram. {c$_{1}$} is the threshold parameter of the color number in the formula and {c$_{1}$} = 0.01, which means if the number of the color in RGB is more then 51, and we can consider it as the RGB dominant color. We define the number of RGB dominant colors (\textit{f2}) as follows:

\begin{equation}
	\begin{aligned}
	f_{2}=\sum_{k=0}^{512} 1\left(h_{k} \geq c_{1} \max _{i} h_{i}\right)
	\end{aligned}
\end{equation}

 The degree of RGB dominant colors (\textit{f3}) indicates that how the dominant color is occupied in the image colors. The formula is as follows:

\begin{equation}
	\begin{aligned}
		f_{3}=\frac{\max _{i} h_{i}}{|I|}
	\end{aligned}
\end{equation}

Similarly, replacing the RGB channel with the HSV channel, we can get the number of HSV dominant colors (\textit{f4}) and the degree of HSV dominant colors (\textit{f5}).

For \textit{f6} and \textit{f7}, we eliminated pixels with saturation elimination values less than 0.2, in other words, we eliminated all light or dark pixels. We then calculate the hue histograms of the remaining pixels with 20 uniform intervals, each occupying 18° sectors of the hue ring. In the hue histogram {H$_{hue}$} = \{{h$_{0}$}, {h$_{1}$}, …, {h$_{19}$}\}, {h$_{i}$} represents the pixel histogram in the i-th interval. The formula for the number of dominant hues is as follows:

\begin{equation}
	\begin{aligned}
		f_{6}=\sum_{i=1}^{20} \mathbf{1}\left(\left|h_{i}\right| \geq c_{2}|I|\right)
	\end{aligned}  
\end{equation}

Similarly, we set {c$_{2}$} = 0.01. The contrast ratio of dominant hues (\textit{f7}) indicates the maximum contrast between the two principal hues in the image, the formula is as follows:

\begin{equation}
	\begin{aligned}
		f_{7}=\max _{i, j}\left\|\left|h_{i}\right|-\left|h_{j}\right|\right\|
	\end{aligned}
\end{equation}

Fig.7 shows four sample images of different color effects, where Fig.7(a) and Fig.7(b) have better color effects, while Fig.7(c) and Fig.7(d) have poor color effect. Table 3 shows the color feature values corresponding to these four images. From the feature values in Table 3, we can see that the images with better color effect generally have more RGB and HSV dominant colors. Besides, the degree and the contrast ratio is higher. So we can find that the images have the better color effects when they have richer colors.

\begin{figure}[ht]
	\centering
	\includegraphics[width=\textwidth]{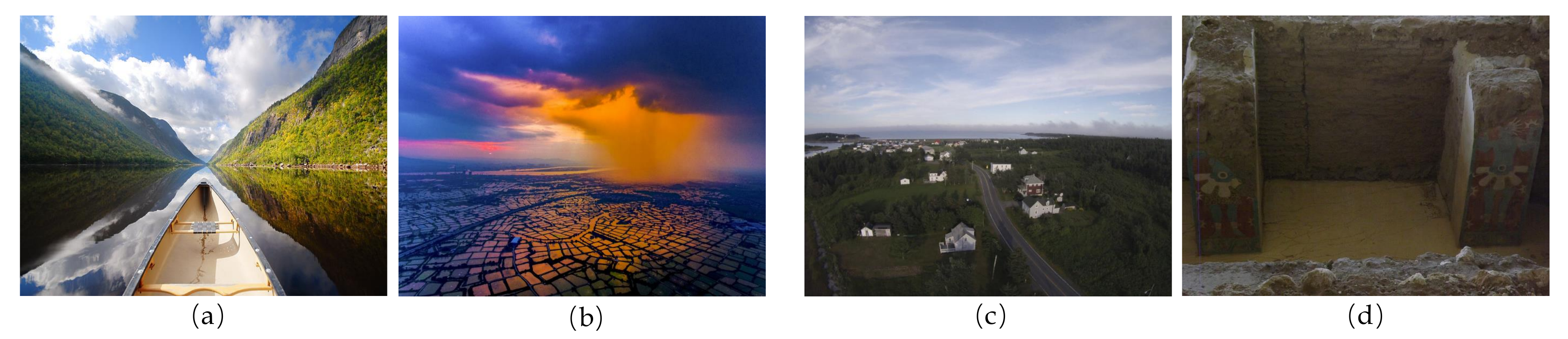}
	\caption{Color feature values for samples}
\end{figure}

\begin{table}[htbp]
	\caption{Calculation results of Color feature sample}
	\centering
	\label{tab:2}
	\begin{tabular}{ccccc}
		\hline\noalign{\smallskip}
		Color feature & 7(a)  & 7(b) & 7(c) & 7(d)  \\
		\noalign{\smallskip}\hline\noalign{\smallskip}
		Weight of color channel \textit{f1} & 1.0 & 1.0 & 1.0 & 1.0  \\
		Number of RGB dominant colors \textit{f2} & 102 & 100 & 19 & 14  \\
		Degree of RGB dominant colors \textit{f3} & 0.087823 & 0.075602 & 0.253429 & 0.228055 \\
		Number of HSV dominant colors \textit{f4} & 220 & 171 & 49 & 39  \\
		Degree of HSV dominant colors \textit{f5} & 0.040854 & 0.052193 & 0.160361 & 0.197352  \\
		Number of dominant hues \textit{f6} & 8 & 10 & 5 & 4  \\
		Contrast ratio of dominant hues \textit{f7} & 162 & 162 & 18 & 18  \\
		\noalign{\smallskip}\hline
	\end{tabular}
\end{table}

\subsection{Composition Features}
We extract the composition features as golden section (\textit{f1}), center (\textit{f2}), slant (\textit{f3}), the triangle (\textit{f4}), guideline (\textit{f5}), rule of thirds (\textit{f6}), symmetry (\textit{f7}), diagonal (\textit{f8}), frame (\textit{f9}) and circle (\textit{f10}).

For the composition features \textit{f1} and \textit{f2}, we first obtain the salient area of the images through salient detection \cite{liu2020dynamic}, and then calculate the distance \textit{d} from the golden section and central points of the image according to the central position of the pixel coordinates in the significance area. After that, we divide the distance \textit{d} by the oblique length of the image to indicate the proximity to the composition. To numerically represent the positive correlation of the compositional feature values, we subtract this value by \textit{1} and represent it as \textit{f1} and \textit{f2}. The formula of the features is as follows:

\begin{equation}
	\begin{aligned}
		f_{1,2}=1-\frac{d}{\sqrt{w^{2}+h^{2}}}
	\end{aligned}
\end{equation}

We obtain the composition feature line of the images by the model in \cite{HanEccv20SemLine} and  calculate \textit{f3}-\textit{f6}. Firstly, we determine the composition feature lines according to the rule of thirds line, symmetrical line, diagonal line and slash line. Then, we respectively calculate the distances between the two endpoints of each composition feature line and the two endpoints of the composition feature line, divide them by the long side of the image and average them. Finally, we subtract this value with 1 to represent the compositional feature calculated values.To better distinguish the accuracy of the values, we set a threshold as 0.3, above which yields a positive value, or 0 otherwise. The calculation formula for \textit{f3}-\textit{f6} is as follows:

\begin{equation}
	\begin{aligned}
		f_{3-6}=1-\frac{d_{1}+d_{2}}{2 \max (w, h)}
	\end{aligned}
\end{equation}

\begin{figure}[htbp]
	\centering
	\includegraphics[width=\textwidth]{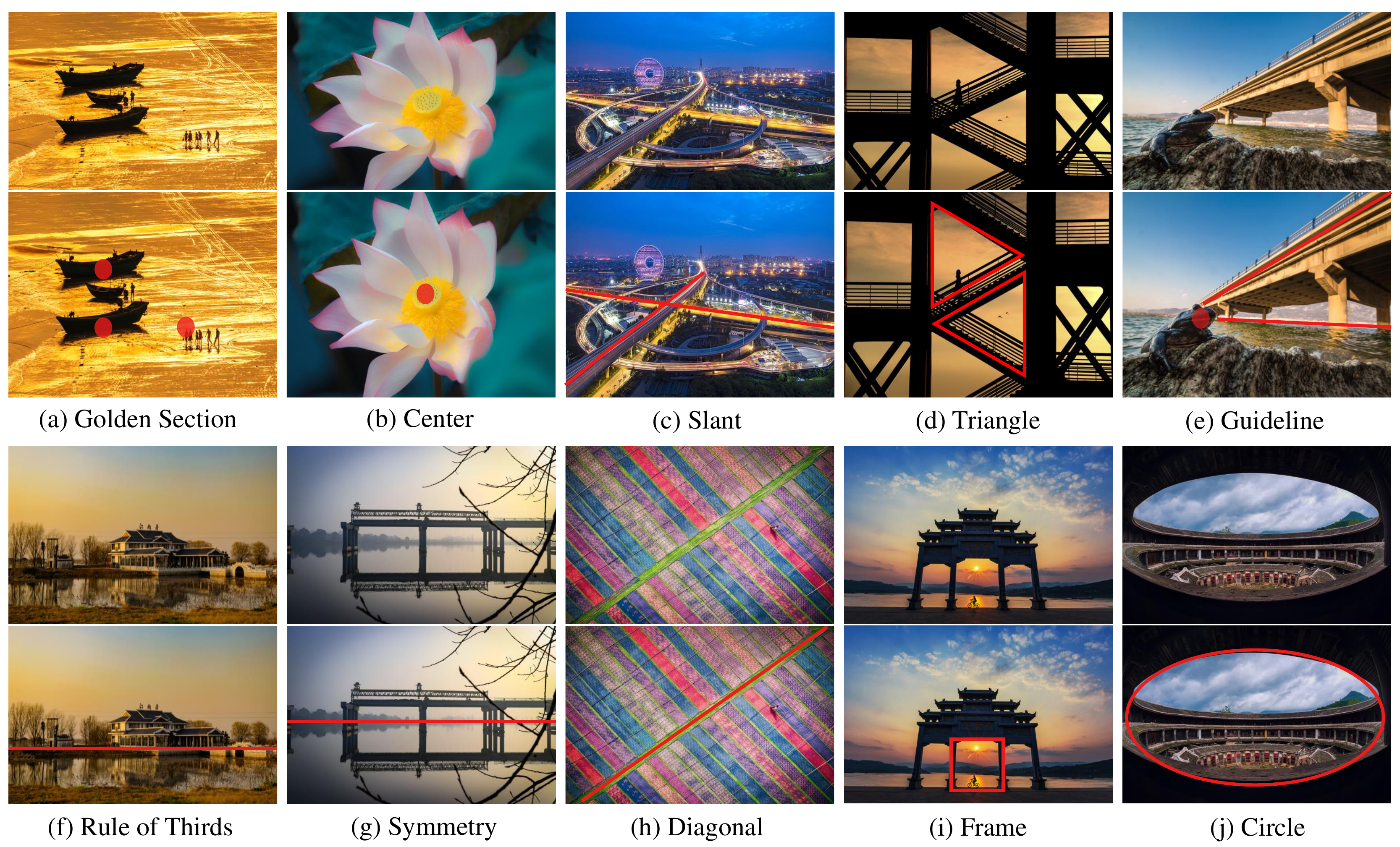}
	\caption{Samples with diffenrent composition effects}
\end{figure}

\begin{table}[htbp]
	\caption{Composition feature values for samples}
	\centering
	\label{tab:1}
	\begin{tabular}{ccccccccccc}
		\hline\noalign{\smallskip}
		Composition feature & 11(a)  & 11(b) & 11(c) & 11(d) & 11(e) & 11(f) & 11(g) & 11(h) & 11(i) & 11(j) \\
		\noalign{\smallskip}\hline\noalign{\smallskip}
		Golden Section \textit{f1} & 0.67 & 0 & 0 & 0 & 0 & 0 & 0 & 0 & 0 & 0   \\
		Center \textit{f2} & 0 & 0.85 & 0 & 0 & 0 & 0.66 & 0.76 & 0 & 0.69 & 0.70 \\
		Slant \textit{f3} & 0 & 0 & 0.48 & 0 & 0.41 & 0 & 0 & 0.71 & 0 & 0 \\
		Triangle \textit{f4} & 0 & 0 & 0 & 0.44 & 0 & 0 & 0 & 0 & 0 & 0 \\
		Guildline \textit{f5} & 0 & 0 & 0 & 0 & 0.71 & 0 & 0 & 0 & 0 & 0 \\
		Rule of Thirds \textit{f6} & 0 & 0 & 0 & 0 & 0.48 & 0.84 & 0 & 0 & 0 & 0 \\
		Symmetry \textit{f7} & 0 & 0 & 0 & 0 & 0 & 0 & 1 & 0 & 0 & 0 \\
		Diagonal \textit{f8} & 0 & 0 & 0 & 0 & 0 & 0 & 0 & 1 & 0 & 0 \\
		Frame \textit{f9} & 0 & 0 & 0 & 0 & 0 & 0 & 0 & 0 & 1 & 0 \\
		Circle \textit{f10} & 0 & 0 & 0 & 0 & 0 & 0 & 0 & 0 & 0 & 1 \\
		\noalign{\smallskip}\hline
	\end{tabular}
\end{table}

We obtained features \textit{f7}-\textit{f9} by calculating whether the intersection positions of the main composition lines are within the range of the images. If there is an intersection of the composition feature line on the image, we will calculate the horizontal angle of the composition feature line. If more than a pair of horizontal angles are complementary, and the distances between the intersections of the composition feature lines are within 0.05max(w, h), \textit{f7} = 1. If the sum of any three composition feature lines and the horizontal surface is 360° and its intersection are within the image, we will consider it as the triangle composition, \textit{f8} = 1. If two composition feature lines are approximately parallel in the horizontal or vertical direction, and another composition feature line is almost vertical to them at the same time, then we will consider it as the frame composition, \textit{f9} = 1. If we detect the image by the hough circle and get the radius of the circle is longer than 0.1min (w, h), we will consider it as the circle composition, \textit{f10} = 1. Otherwise all the above features \textit{f7}-\textit{f10} are 0.

Fig.8 shows ten sample images of different composition effects. Table 4 shows the composition feature values corresponding to these ten images. From the feature values in Table 4, we can see that the images with one or more than one composition features have better composition effects.

\subsection{Feature fusion}
The network structure in Fig.3 shows that the yellow part is the external attribute feature {f$_{external}$}(4 for light features, 7 for color features or 10 for composition features). We concat these features with the aesthetic high-level features extracted from the backbone network by the method mentioned in Section 4.2. The experiment results in Section 6 verify the method of the feature fusion.

\section{Experiment}
\subsection{Training details}
We set the classification batch size to be 32, the regression batch size to be 64 and the learning rate to be 0.0001. We use Adam as the optimizer; betas are set as (0.98, 0.999); weight decay is set as 0.0001. If the accuracy rate of classification is not improved in two consecutive rounds, or the regression loss is not decreased in two consecutive rounds, the learning rate will multiply by 0.5. Our running environment is in MindSpore \cite{mindspore} 1.6.0 and Nvidia TITAN XPs. We divide the dataset into three sets; the ratio of training set and validation set and testing set is 8:1:1.

\subsection{Analysis of results}

This paper uses several indicators to measure training results:

MSE mean square error represents the estimating errors of the difference between the predicted results and the true values; the formula is:

\begin{equation}
	MSE=\frac{1}{N}\sum_{i}{(r_i-\widehat{r_i})}^2
\end{equation}

SROCC (Spearman rankorder correlation coefficient, Spearman rank correlation coefficient) represents the correlation between the predicted result and the true value and the formula is:

\begin{equation}
	SROCC=1-\frac{6\sum_{i}{(r_i-\widehat{r_i})}^2}{N^3-N}
\end{equation}

The accuracy of the two-classification indicates whether the predicted score and the real score are consistent when the boundary is 5. The accuracy of two-classification indicates the most basic accuracy of classification; the formula is:
 
\begin{equation}
	ACCURACY=\frac{TP+TN}{P+N}
\end{equation}

The positive and negative accuracy indicates whether the absolute value of the error is within 1 point between the predicted score and the real score. If the absolute value of the error is within 1 point, the result can be considered accurate. The formula is:

\begin{equation}
	ACCURACY_{\left|error\right|\le1}=\frac{N_{\left|error\right|\le1}}{N}
\end{equation}
\begin{table}[htbp]
	\caption{Feature fussion experimental results}
	\centering
	\label{tab:3}
	\begin{tabular}{cccccc}
		\hline\noalign{\smallskip}
		Methods & Attributes & MSE & SROCC & Acc & $Acc_{\left|error\right|\le1}$  \\
		\noalign{\smallskip}\hline\noalign{\smallskip}
		& Color & 0.008660 & 0.6863 & 76.17\% & 73.68\%  \\
		Baseline (Mindspore) & Light & 0.011266 & 0.6939 & 77.72\% & 68.29\%  \\
		& Composition & 0.009491 & 0.6915 & 77.41\% & 70.57\%  \\
		 \noalign{\smallskip}\hline\noalign{\smallskip}
		 \textbf{Basebone +} & \textbf{Color} & \textbf{0.008307} & \textbf{0.7087} & \textbf{77.72\%} & \textbf{73.69\%}  \\
		 \textbf{Feature fusion} & \textbf{Light} & \textbf{0.010675} & \textbf{0.7126} & \textbf{79.27\%} & \textbf{69.53\%}  \\
		 \textbf{(Mindspore)} & \textbf{Composition} & \textbf{0.009285} & \textbf{0.7013} & \textbf{79.79\%} & \textbf{73.47\%} \\
		\noalign{\smallskip}\hline
	\end{tabular}
\end{table}

\begin{table}[htbp]
	\caption{Soft loss experimental results}
	\centering
	\label{tab:3}
	\begin{tabular}{cccccc}
		\hline\noalign{\smallskip}
		Methods & Attributes & MSE & SROCC & Acc & $Acc_{\left|error\right|\le1}$  \\
		\noalign{\smallskip}\hline\noalign{\smallskip}
		Baseline (Mindspore) & Score & 0.012940 & 0.8424 & 85.78\% & 64.99\%  \\
		\textbf{Baseline + Soft loss (Mindspore)} & \textbf{Score} & \textbf{0.011315} & \textbf{0.8604} & \textbf{88.08\%} & \textbf{68.72\%} \\
		\noalign{\smallskip}\hline
	\end{tabular}
\end{table}
We use EfficientNet-B0 as the baseline and conducte several experiments as follows: for feature fusion, Table 5 showed that external light, color, and composition attribute features could return to a more accurate assessment after fusion. For overall score regression, we used the soft loss mentioned in Section 4.2, and Table 6 showed that the model with soft loss had better performances in all indicators.
\begin{table}[htbp]
	\caption{The comparison experiment for \textit{$\lambda$}}
	\centering
	\label{tab:3}
	\begin{tabular}{ccccc}
		\hline\noalign{\smallskip}
		$\lambda$ & MSE & SROCC & Acc & $Acc_{\left|error\right|\le1}$\\
		\noalign{\smallskip}\hline\noalign{\smallskip}
		0 & 0.011408 & 0.8591 & 88.02\% & 68.31\%\\
		\textbf{0.1} & \textbf{0.011315} & \textbf{0.8604} & \textbf{88.08\%} & \textbf{68.72\%} \\
		0.2  & 0.011318 & 0.8603 & 87.88\% & 68.44\%\\
		0.3  & 0.011402 & 0.8595 & 87.81\% & 68.58\%\\
		0.4  & 0.011437 & 0.8591 & 87.74\% & 68.45\%\\
		\noalign{\smallskip}\hline
	\end{tabular}
\end{table}
We did a comparison experiment in 40 epoches of training. Table 7 showes that the presence of \textit{$\lambda$} does improve the accuracy of the regression, when \textit{$\lambda$} = 0.1, so we set \textit{$\lambda$} to 0.1.

Besides, we selected some typical samples and used this method to score the images, including the overall aesthetic score and the scores of the three aesthetic attributes. Among them, S represents the overall score, C represents the color score, L represents the light score, and CM represents the composition score, as shown in Fig.9:

\begin{figure*}[btp!]
	\centering
	\includegraphics[width=\linewidth]{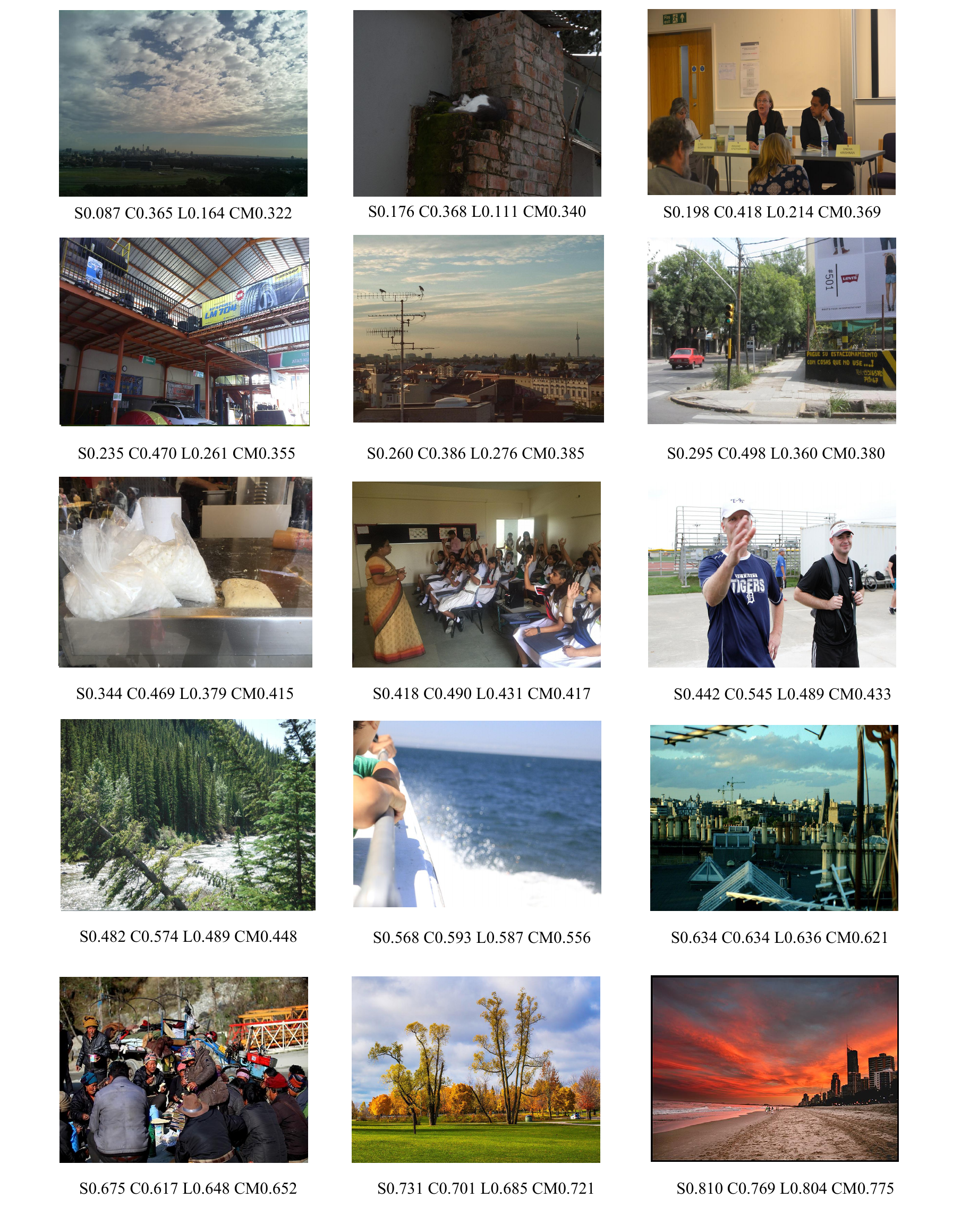}
	\caption{Test samples on the testing set  with scores from low to high. S represents the overall score, C represents the color score, L represents the light score, and CM represents the composition score}
	\label{fig:10}  
\end{figure*}

\section{Summarize}

It is challenging to construct a new dataset in the image aesthetic quality assessment. Traditional datasets have the problem of few data and limited attribute labels. By mixing and filtering massive datasets and designing external attribute features, we get a new dataset called aesthetic mixed dataset with attributes(AMD-A) with a more reasonable distribution. Besides, we propose a model with multitasking network including one classification sub-network, three attribute sub-network and one regression sub-network. This is an innovative exploration of the training method for the numerical assessment of image aesthetics. Moreover, we design and use the external attribute feature fusion to improve the regressing of aesthetic attributes. According to the idea of the teacher-student network, we use the classification sub-network to guide the regression sub-network through high-level features by the soft loss. The experimental results proved that our model is more accurate than the traditional deep learning network regression model, and it improves the prediction of aesthetic attributes and overall scores. 

\section*{Acknowledgements}
This work is partially supported by the National Natural Science Foundation of China (62072014\& 62106118), the CAAI-Huawei MindSpore Open Fund (CAAIXSJLJJ-2021-022A), the Open Fund Project of the State Key Laboratory of Complex System Management and Control (2022111), the Project of Philosophy and Social Science Research, Ministry of Education of China (No.20YJC760115), and the Advanced Discipline Construction Project of Beijing Universities (20210051Z0401).

We gratefully acknowledge the support of MindSpore, CANN (Compute Architecture for Neural Networks), Ascend AI Processor and Zhongyuan AI Computing Center used for this research.

\bibliographystyle{ACM-Reference-Format}
\bibliography{sample-base}


\begin{thebibliography}{42}


\ifx \showCODEN    \undefined \def \showCODEN     #1{\unskip}     \fi
\ifx \showDOI      \undefined \def \showDOI       #1{#1}\fi
\ifx \showISBNx    \undefined \def \showISBNx     #1{\unskip}     \fi
\ifx \showISBNxiii \undefined \def \showISBNxiii  #1{\unskip}     \fi
\ifx \showISSN     \undefined \def \showISSN      #1{\unskip}     \fi
\ifx \showLCCN     \undefined \def \showLCCN      #1{\unskip}     \fi
\ifx \shownote     \undefined \def \shownote      #1{#1}          \fi
\ifx \showarticletitle \undefined \def \showarticletitle #1{#1}   \fi
\ifx \showURL      \undefined \def \showURL       {\relax}        \fi
\providecommand\bibfield[2]{#2}
\providecommand\bibinfo[2]{#2}
\providecommand\natexlab[1]{#1}
\providecommand\showeprint[2][]{arXiv:#2}

\bibitem[\protect\citeauthoryear{Azimi, Zhang, Zhou, Navalpakkam, Mao, and
  Fern}{Azimi et~al\mbox{.}}{2012}]%
        {2012The}
\bibfield{author}{\bibinfo{person}{Javad Azimi}, \bibinfo{person}{Ruofei
  Zhang}, \bibinfo{person}{Yang Zhou}, \bibinfo{person}{Vidhya Navalpakkam},
  \bibinfo{person}{Jianchang Mao}, {and} \bibinfo{person}{Xiaoli Fern}.}
  \bibinfo{year}{2012}\natexlab{}.
\newblock \showarticletitle{The impact of visual appearance on user response in
  online display advertising}. In \bibinfo{booktitle}{\emph{proceedings of the
  21st international conference on World Wide Web}}. \bibinfo{pages}{457--458}.
\newblock


\bibitem[\protect\citeauthoryear{Bell, Zitnick, Bala, and Girshick}{Bell
  et~al\mbox{.}}{2016}]%
        {2016Inside}
\bibfield{author}{\bibinfo{person}{Sean Bell}, \bibinfo{person}{C~Lawrence
  Zitnick}, \bibinfo{person}{Kavita Bala}, {and} \bibinfo{person}{Ross
  Girshick}.} \bibinfo{year}{2016}\natexlab{}.
\newblock \showarticletitle{Inside-outside net: Detecting objects in context
  with skip pooling and recurrent neural networks}. In
  \bibinfo{booktitle}{\emph{Proceedings of the IEEE conference on computer
  vision and pattern recognition}}. \bibinfo{pages}{2874--2883}.
\newblock


\bibitem[\protect\citeauthoryear{Brachmann and Redies}{Brachmann and
  Redies}{2017}]%
        {2017Computational}
\bibfield{author}{\bibinfo{person}{Anselm Brachmann} {and}
  \bibinfo{person}{Christoph Redies}.} \bibinfo{year}{2017}\natexlab{}.
\newblock \showarticletitle{Computational and experimental approaches to visual
  aesthetics}.
\newblock \bibinfo{journal}{\emph{Frontiers in computational neuroscience}}
  \bibinfo{volume}{11} (\bibinfo{year}{2017}), \bibinfo{pages}{102}.
\newblock


\bibitem[\protect\citeauthoryear{Chang, Lu, and Chen}{Chang
  et~al\mbox{.}}{2017}]%
        {2017Aesthetic}
\bibfield{author}{\bibinfo{person}{Kuang-Yu Chang}, \bibinfo{person}{Kung-Hung
  Lu}, {and} \bibinfo{person}{Chu-Song Chen}.} \bibinfo{year}{2017}\natexlab{}.
\newblock \showarticletitle{Aesthetic critiques generation for photos}. In
  \bibinfo{booktitle}{\emph{Proceedings of the IEEE international conference on
  computer vision}}. \bibinfo{pages}{3514--3523}.
\newblock


\bibitem[\protect\citeauthoryear{Datta, Joshi, Li, and Wang}{Datta
  et~al\mbox{.}}{2006}]%
        {2008STUDYING}
\bibfield{author}{\bibinfo{person}{Ritendra Datta}, \bibinfo{person}{Dhiraj
  Joshi}, \bibinfo{person}{Jia Li}, {and} \bibinfo{person}{James~Z Wang}.}
  \bibinfo{year}{2006}\natexlab{}.
\newblock \showarticletitle{Studying aesthetics in photographic images using a
  computational approach}. In \bibinfo{booktitle}{\emph{European conference on
  computer vision}}. Springer, \bibinfo{pages}{288--301}.
\newblock


\bibitem[\protect\citeauthoryear{Deng, Loy, and Tang}{Deng
  et~al\mbox{.}}{2018}]%
        {2017Aestheticd}
\bibfield{author}{\bibinfo{person}{Yubin Deng}, \bibinfo{person}{Chen~Change
  Loy}, {and} \bibinfo{person}{Xiaoou Tang}.} \bibinfo{year}{2018}\natexlab{}.
\newblock \showarticletitle{Aesthetic-driven image enhancement by adversarial
  learning}. In \bibinfo{booktitle}{\emph{Proceedings of the 26th ACM
  international conference on Multimedia}}. \bibinfo{pages}{870--878}.
\newblock


\bibitem[\protect\citeauthoryear{Fang, Zhu, Zeng, Ma, and Wang}{Fang
  et~al\mbox{.}}{2020}]%
        {2020Perceptual}
\bibfield{author}{\bibinfo{person}{Yuming Fang}, \bibinfo{person}{Hanwei Zhu},
  \bibinfo{person}{Yan Zeng}, \bibinfo{person}{Kede Ma}, {and}
  \bibinfo{person}{Zhou Wang}.} \bibinfo{year}{2020}\natexlab{}.
\newblock \showarticletitle{Perceptual quality assessment of smartphone
  photography}. In \bibinfo{booktitle}{\emph{Proceedings of the IEEE/CVF
  Conference on Computer Vision and Pattern Recognition}}.
  \bibinfo{pages}{3677--3686}.
\newblock


\bibitem[\protect\citeauthoryear{Han, Zhao, Xu, and Cheng}{Han
  et~al\mbox{.}}{2020}]%
        {HanEccv20SemLine}
\bibfield{author}{\bibinfo{person}{Qi Han}, \bibinfo{person}{Kai Zhao},
  \bibinfo{person}{Jun Xu}, {and} \bibinfo{person}{Ming-Ming Cheng}.}
  \bibinfo{year}{2020}\natexlab{}.
\newblock \showarticletitle{Deep Hough Transform for Semantic Line Detection}.
  In \bibinfo{booktitle}{\emph{European Conference on Computer Vision (ECCV)}}.
  \bibinfo{pages}{249--265}.
\newblock
\urldef\tempurl%
\url{https://doi.org/10.1007/978-3-030-58545-7_15}
\showDOI{\tempurl}


\bibitem[\protect\citeauthoryear{Hinton, Vinyals, Dean, et~al\mbox{.}}{Hinton
  et~al\mbox{.}}{2015}]%
        {2015Distilling}
\bibfield{author}{\bibinfo{person}{Geoffrey Hinton}, \bibinfo{person}{Oriol
  Vinyals}, \bibinfo{person}{Jeff Dean}, {et~al\mbox{.}}}
  \bibinfo{year}{2015}\natexlab{}.
\newblock \showarticletitle{Distilling the knowledge in a neural network}.
\newblock \bibinfo{journal}{\emph{arXiv preprint arXiv:1503.02531}}
  \bibinfo{volume}{2}, \bibinfo{number}{7} (\bibinfo{year}{2015}).
\newblock


\bibitem[\protect\citeauthoryear{Huang, Liu, Van Der~Maaten, and
  Weinberger}{Huang et~al\mbox{.}}{2017}]%
        {2016Densely}
\bibfield{author}{\bibinfo{person}{Gao Huang}, \bibinfo{person}{Zhuang Liu},
  \bibinfo{person}{Laurens Van Der~Maaten}, {and} \bibinfo{person}{Kilian~Q
  Weinberger}.} \bibinfo{year}{2017}\natexlab{}.
\newblock \showarticletitle{Densely connected convolutional networks}. In
  \bibinfo{booktitle}{\emph{Proceedings of the IEEE conference on computer
  vision and pattern recognition}}. \bibinfo{pages}{4700--4708}.
\newblock


\bibitem[\protect\citeauthoryear{Jin, Wu, Zhou, Zhao, Zhang, Li, and Ge}{Jin
  et~al\mbox{.}}{2018}]%
        {2018Predicting}
\bibfield{author}{\bibinfo{person}{Xin Jin}, \bibinfo{person}{Le Wu},
  \bibinfo{person}{Xinghui Zhou}, \bibinfo{person}{Geng Zhao},
  \bibinfo{person}{Xiaokun Zhang}, \bibinfo{person}{Xiaodong Li}, {and}
  \bibinfo{person}{Shiming Ge}.} \bibinfo{year}{2018}\natexlab{}.
\newblock \showarticletitle{Predicting aesthetic radar map using a hierarchical
  multi-task network}. In \bibinfo{booktitle}{\emph{Chinese Conference on
  Pattern Recognition and Computer Vision (PRCV)}}. Springer,
  \bibinfo{pages}{41--50}.
\newblock


\bibitem[\protect\citeauthoryear{Jin, Zhou, Li, Zhang, Sun, Li, and Liu}{Jin
  et~al\mbox{.}}{2019}]%
        {2019Incremental}
\bibfield{author}{\bibinfo{person}{Xin Jin}, \bibinfo{person}{Xinghui Zhou},
  \bibinfo{person}{Xiaodong Li}, \bibinfo{person}{Xiaokun Zhang},
  \bibinfo{person}{Hongbo Sun}, \bibinfo{person}{Xiqiao Li}, {and}
  \bibinfo{person}{Ruijun Liu}.} \bibinfo{year}{2019}\natexlab{}.
\newblock \showarticletitle{Incremental Learning of Multi-Tasking Networks for
  Aesthetic Radar Map Prediction}.
\newblock \bibinfo{journal}{\emph{IEEE Access}}  \bibinfo{volume}{7}
  (\bibinfo{year}{2019}), \bibinfo{pages}{183647--183655}.
\newblock


\bibitem[\protect\citeauthoryear{Kairanbay, See, and Wong}{Kairanbay
  et~al\mbox{.}}{2019}]%
        {kairanbay2019beauty}
\bibfield{author}{\bibinfo{person}{Magzhan Kairanbay}, \bibinfo{person}{John
  See}, {and} \bibinfo{person}{Lai-Kuan Wong}.}
  \bibinfo{year}{2019}\natexlab{}.
\newblock \showarticletitle{Beauty is in the eye of the beholder:
  Demographically oriented analysis of aesthetics in photographs}.
\newblock \bibinfo{journal}{\emph{ACM Transactions on Multimedia Computing,
  Communications, and Applications (TOMM)}} \bibinfo{volume}{15},
  \bibinfo{number}{2s} (\bibinfo{year}{2019}), \bibinfo{pages}{1--21}.
\newblock


\bibitem[\protect\citeauthoryear{Kang, Valenzise, and Dufaux}{Kang
  et~al\mbox{.}}{2020}]%
        {kang2020eva}
\bibfield{author}{\bibinfo{person}{Chen Kang}, \bibinfo{person}{Giuseppe
  Valenzise}, {and} \bibinfo{person}{Fr{\'e}d{\'e}ric Dufaux}.}
  \bibinfo{year}{2020}\natexlab{}.
\newblock \showarticletitle{EVA: An Explainable Visual Aesthetics Dataset}. In
  \bibinfo{booktitle}{\emph{Joint Workshop on Aesthetic and Technical Quality
  Assessment of Multimedia and Media Analytics for Societal Trends}}.
  \bibinfo{pages}{5--13}.
\newblock


\bibitem[\protect\citeauthoryear{Kao, Wang, and Huang}{Kao
  et~al\mbox{.}}{2015}]%
        {2015Visual}
\bibfield{author}{\bibinfo{person}{Yueying Kao}, \bibinfo{person}{Chong Wang},
  {and} \bibinfo{person}{Kaiqi Huang}.} \bibinfo{year}{2015}\natexlab{}.
\newblock \showarticletitle{Visual aesthetic quality assessment with a
  regression model}. In \bibinfo{booktitle}{\emph{2015 IEEE International
  Conference on Image Processing (ICIP)}}. IEEE, \bibinfo{pages}{1583--1587}.
\newblock


\bibitem[\protect\citeauthoryear{Ke, Tang, and Jing}{Ke et~al\mbox{.}}{2006}]%
        {1640788}
\bibfield{author}{\bibinfo{person}{Yan Ke}, \bibinfo{person}{Xiaoou Tang},
  {and} \bibinfo{person}{Feng Jing}.} \bibinfo{year}{2006}\natexlab{}.
\newblock \showarticletitle{The design of high-level features for photo quality
  assessment}. In \bibinfo{booktitle}{\emph{2006 IEEE Computer Society
  Conference on Computer Vision and Pattern Recognition (CVPR'06)}},
  Vol.~\bibinfo{volume}{1}. IEEE, \bibinfo{pages}{419--426}.
\newblock


\bibitem[\protect\citeauthoryear{Kong, Shen, Lin, Mech, and Fowlkes}{Kong
  et~al\mbox{.}}{2016a}]%
        {2016Photo}
\bibfield{author}{\bibinfo{person}{Shu Kong}, \bibinfo{person}{Xiaohui Shen},
  \bibinfo{person}{Zhe Lin}, \bibinfo{person}{Radomir Mech}, {and}
  \bibinfo{person}{Charless Fowlkes}.} \bibinfo{year}{2016}\natexlab{a}.
\newblock \showarticletitle{Photo aesthetics ranking network with attributes
  and content adaptation}. In \bibinfo{booktitle}{\emph{European conference on
  computer vision}}. Springer, \bibinfo{pages}{662--679}.
\newblock


\bibitem[\protect\citeauthoryear{Kong, Yao, Chen, and Sun}{Kong
  et~al\mbox{.}}{2016b}]%
        {2016HyperNet}
\bibfield{author}{\bibinfo{person}{Tao Kong}, \bibinfo{person}{Anbang Yao},
  \bibinfo{person}{Yurong Chen}, {and} \bibinfo{person}{Fuchun Sun}.}
  \bibinfo{year}{2016}\natexlab{b}.
\newblock \showarticletitle{Hypernet: Towards accurate region proposal
  generation and joint object detection}. In
  \bibinfo{booktitle}{\emph{Proceedings of the IEEE conference on computer
  vision and pattern recognition}}. \bibinfo{pages}{845--853}.
\newblock


\bibitem[\protect\citeauthoryear{Liang, Lin, Jin, Xie, and Li}{Liang
  et~al\mbox{.}}{2018}]%
        {2018SCUT}
\bibfield{author}{\bibinfo{person}{Lingyu Liang}, \bibinfo{person}{Luojun Lin},
  \bibinfo{person}{Lianwen Jin}, \bibinfo{person}{Duorui Xie}, {and}
  \bibinfo{person}{Mengru Li}.} \bibinfo{year}{2018}\natexlab{}.
\newblock \showarticletitle{SCUT-FBP5500: a diverse benchmark dataset for
  multi-paradigm facial beauty prediction}. In \bibinfo{booktitle}{\emph{2018
  24th International Conference on Pattern Recognition (ICPR)}}. IEEE,
  \bibinfo{pages}{1598--1603}.
\newblock


\bibitem[\protect\citeauthoryear{Lin, Doll{\'a}r, Girshick, He, Hariharan, and
  Belongie}{Lin et~al\mbox{.}}{2017}]%
        {2017Feature}
\bibfield{author}{\bibinfo{person}{Tsung-Yi Lin}, \bibinfo{person}{Piotr
  Doll{\'a}r}, \bibinfo{person}{Ross Girshick}, \bibinfo{person}{Kaiming He},
  \bibinfo{person}{Bharath Hariharan}, {and} \bibinfo{person}{Serge Belongie}.}
  \bibinfo{year}{2017}\natexlab{}.
\newblock \showarticletitle{Feature pyramid networks for object detection}. In
  \bibinfo{booktitle}{\emph{Proceedings of the IEEE conference on computer
  vision and pattern recognition}}. \bibinfo{pages}{2117--2125}.
\newblock


\bibitem[\protect\citeauthoryear{Liu, Hou, and Cheng}{Liu
  et~al\mbox{.}}{2020}]%
        {liu2020dynamic}
\bibfield{author}{\bibinfo{person}{Jiang-Jiang Liu}, \bibinfo{person}{Qibin
  Hou}, {and} \bibinfo{person}{Ming-Ming Cheng}.}
  \bibinfo{year}{2020}\natexlab{}.
\newblock \showarticletitle{Dynamic feature integration for simultaneous
  detection of salient object, edge, and skeleton}.
\newblock \bibinfo{journal}{\emph{IEEE Transactions on Image Processing}}
  \bibinfo{volume}{29} (\bibinfo{year}{2020}), \bibinfo{pages}{8652--8667}.
\newblock


\bibitem[\protect\citeauthoryear{Liu, Anguelov, Erhan, Szegedy, Reed, Fu, and
  Berg}{Liu et~al\mbox{.}}{2016}]%
        {2016SSD}
\bibfield{author}{\bibinfo{person}{Wei Liu}, \bibinfo{person}{Dragomir
  Anguelov}, \bibinfo{person}{Dumitru Erhan}, \bibinfo{person}{Christian
  Szegedy}, \bibinfo{person}{Scott Reed}, \bibinfo{person}{Cheng-Yang Fu},
  {and} \bibinfo{person}{Alexander~C Berg}.} \bibinfo{year}{2016}\natexlab{}.
\newblock \showarticletitle{Ssd: Single shot multibox detector}. In
  \bibinfo{booktitle}{\emph{European conference on computer vision}}. Springer,
  \bibinfo{pages}{21--37}.
\newblock


\bibitem[\protect\citeauthoryear{Luo and Tang}{Luo and Tang}{2008}]%
        {2008Photo}
\bibfield{author}{\bibinfo{person}{Yiwen Luo} {and} \bibinfo{person}{Xiaoou
  Tang}.} \bibinfo{year}{2008}\natexlab{}.
\newblock \showarticletitle{Photo and video quality evaluation: Focusing on the
  subject}. In \bibinfo{booktitle}{\emph{European conference on computer
  vision}}. Springer, \bibinfo{pages}{386--399}.
\newblock


\bibitem[\protect\citeauthoryear{Ma, Liu, and Wen~Chen}{Ma
  et~al\mbox{.}}{2017}]%
        {2017A}
\bibfield{author}{\bibinfo{person}{Shuang Ma}, \bibinfo{person}{Jing Liu},
  {and} \bibinfo{person}{Chang Wen~Chen}.} \bibinfo{year}{2017}\natexlab{}.
\newblock \showarticletitle{A-lamp: Adaptive layout-aware multi-patch deep
  convolutional neural network for photo aesthetic assessment}. In
  \bibinfo{booktitle}{\emph{Proceedings of the IEEE conference on computer
  vision and pattern recognition}}. \bibinfo{pages}{4535--4544}.
\newblock


\bibitem[\protect\citeauthoryear{Marchesotti, Perronnin, Larlus, and
  Csurka}{Marchesotti et~al\mbox{.}}{2011a}]%
        {2011Assessing}
\bibfield{author}{\bibinfo{person}{Luca Marchesotti}, \bibinfo{person}{Florent
  Perronnin}, \bibinfo{person}{Diane Larlus}, {and} \bibinfo{person}{Gabriela
  Csurka}.} \bibinfo{year}{2011}\natexlab{a}.
\newblock \showarticletitle{Assessing the aesthetic quality of photographs
  using generic image descriptors}. In \bibinfo{booktitle}{\emph{2011
  international conference on computer vision}}. IEEE,
  \bibinfo{pages}{1784--1791}.
\newblock


\bibitem[\protect\citeauthoryear{Marchesotti, Perronnin, Larlus, and
  Csurka}{Marchesotti et~al\mbox{.}}{2011b}]%
        {6126444}
\bibfield{author}{\bibinfo{person}{Luca Marchesotti}, \bibinfo{person}{Florent
  Perronnin}, \bibinfo{person}{Diane Larlus}, {and} \bibinfo{person}{Gabriela
  Csurka}.} \bibinfo{year}{2011}\natexlab{b}.
\newblock \showarticletitle{Assessing the aesthetic quality of photographs
  using generic image descriptors}. In \bibinfo{booktitle}{\emph{2011
  international conference on computer vision}}. IEEE,
  \bibinfo{pages}{1784--1791}.
\newblock


\bibitem[\protect\citeauthoryear{MindSpore}{MindSpore}{2022}]%
        {mindspore}
\bibfield{author}{\bibinfo{person}{MindSpore}.}
  \bibinfo{year}{2022}\natexlab{}.
\newblock \bibinfo{howpublished}{\url{https://www.mindspore.cn/}}.
\newblock


\bibitem[\protect\citeauthoryear{Murray, Marchesotti, and Perronnin}{Murray
  et~al\mbox{.}}{2012}]%
        {2012AVA}
\bibfield{author}{\bibinfo{person}{Naila Murray}, \bibinfo{person}{Luca
  Marchesotti}, {and} \bibinfo{person}{Florent Perronnin}.}
  \bibinfo{year}{2012}\natexlab{}.
\newblock \showarticletitle{AVA: A large-scale database for aesthetic visual
  analysis}. In \bibinfo{booktitle}{\emph{2012 IEEE conference on computer
  vision and pattern recognition}}. IEEE, \bibinfo{pages}{2408--2415}.
\newblock


\bibitem[\protect\citeauthoryear{Nishiyama, Okabe, Sato, and Sato}{Nishiyama
  et~al\mbox{.}}{2011}]%
        {2011Aesthetic}
\bibfield{author}{\bibinfo{person}{Masashi Nishiyama},
  \bibinfo{person}{Takahiro Okabe}, \bibinfo{person}{Imari Sato}, {and}
  \bibinfo{person}{Yoichi Sato}.} \bibinfo{year}{2011}\natexlab{}.
\newblock \showarticletitle{Aesthetic quality classification of photographs
  based on color harmony}. In \bibinfo{booktitle}{\emph{CVPR 2011}}. IEEE,
  \bibinfo{pages}{33--40}.
\newblock


\bibitem[\protect\citeauthoryear{Qilong~Wang and Hu.}{Qilong~Wang and
  Hu.}{2020}]%
        {2019ECA}
\bibfield{author}{\bibinfo{person}{Pengfei Zhu Peihua Li Wangmeng~Zuo
  Qilong~Wang, Banggu~Wu} {and} \bibinfo{person}{Qinghua Hu.}}
  \bibinfo{year}{2020}\natexlab{}.
\newblock \showarticletitle{ECA-Net: Efficient Channel Attention for Deep
  Convolutional Neural Networks}.
\newblock  (\bibinfo{year}{2020}), \bibinfo{pages}{11531–--11539}.
\newblock


\bibitem[\protect\citeauthoryear{She, Lai, Yi, and Xu}{She
  et~al\mbox{.}}{2021}]%
        {2021Hierarchical}
\bibfield{author}{\bibinfo{person}{Dongyu She}, \bibinfo{person}{Yu-Kun Lai},
  \bibinfo{person}{Gaoxiong Yi}, {and} \bibinfo{person}{Kun Xu}.}
  \bibinfo{year}{2021}\natexlab{}.
\newblock \showarticletitle{Hierarchical layout-aware graph convolutional
  network for unified aesthetics assessment}. In
  \bibinfo{booktitle}{\emph{Proceedings of the IEEE/CVF Conference on Computer
  Vision and Pattern Recognition}}. \bibinfo{pages}{8475--8484}.
\newblock


\bibitem[\protect\citeauthoryear{Sperry, Gazzaniga, and Bogen}{Sperry
  et~al\mbox{.}}{1969}]%
        {1969Interhemispheric}
\bibfield{author}{\bibinfo{person}{R.~W. Sperry}, \bibinfo{person}{M.~S.
  Gazzaniga}, {and} \bibinfo{person}{J.~E. Bogen}.}
  \bibinfo{year}{1969}\natexlab{}.
\newblock \showarticletitle{Interhemispheric relationships: the neocortical
  commissures; syndromes of hemisphere disconnection}.
\newblock \bibinfo{journal}{\emph{Handbook of Clinical Neurology}}
  \bibinfo{volume}{4} (\bibinfo{year}{1969}), \bibinfo{pages}{145--153}.
\newblock


\bibitem[\protect\citeauthoryear{Tan and Le}{Tan and Le}{2019}]%
        {2019EfficientNet}
\bibfield{author}{\bibinfo{person}{Mingxing Tan} {and} \bibinfo{person}{Quoc
  Le}.} \bibinfo{year}{2019}\natexlab{}.
\newblock \showarticletitle{Efficientnet: Rethinking model scaling for
  convolutional neural networks}. In \bibinfo{booktitle}{\emph{International
  conference on machine learning}}. PMLR, \bibinfo{pages}{6105--6114}.
\newblock


\bibitem[\protect\citeauthoryear{Tong, Li, Zhang, He, and Zhang}{Tong
  et~al\mbox{.}}{2004}]%
        {2004Classification}
\bibfield{author}{\bibinfo{person}{Hanghang Tong}, \bibinfo{person}{Mingjing
  Li}, \bibinfo{person}{Hong-Jiang Zhang}, \bibinfo{person}{Jingrui He}, {and}
  \bibinfo{person}{Changshui Zhang}.} \bibinfo{year}{2004}\natexlab{}.
\newblock \showarticletitle{Classification of digital photos taken by
  photographers or home users}. In \bibinfo{booktitle}{\emph{Pacific-Rim
  Conference on Multimedia}}. Springer, \bibinfo{pages}{198--205}.
\newblock


\bibitem[\protect\citeauthoryear{Wang, Zhang, Fortino, Guan, Liu, and
  Song}{Wang et~al\mbox{.}}{2022a}]%
        {wang2022software}
\bibfield{author}{\bibinfo{person}{Ranran Wang}, \bibinfo{person}{Yin Zhang},
  \bibinfo{person}{Giancarlo Fortino}, \bibinfo{person}{Qingxu Guan},
  \bibinfo{person}{Jiangchuan Liu}, {and} \bibinfo{person}{Jeungeun Song}.}
  \bibinfo{year}{2022}\natexlab{a}.
\newblock \showarticletitle{Software Escalation Prediction Based on Deep
  Learning in the Cognitive Internet of Vehicles}.
\newblock \bibinfo{journal}{\emph{IEEE TRANSACTIONS ON INTELLIGENT
  TRANSPORTATION SYSTEMS}} (\bibinfo{year}{2022}).
\newblock


\bibitem[\protect\citeauthoryear{Wang, Zhang, Peng, Fortino, and Ho}{Wang
  et~al\mbox{.}}{2022b}]%
        {wang2022time}
\bibfield{author}{\bibinfo{person}{Ranran Wang}, \bibinfo{person}{Yin Zhang},
  \bibinfo{person}{Limei Peng}, \bibinfo{person}{Giancarlo Fortino}, {and}
  \bibinfo{person}{Pin-Han Ho}.} \bibinfo{year}{2022}\natexlab{b}.
\newblock \showarticletitle{Time-varying-aware network traffic prediction via
  deep learning in IIoT}.
\newblock \bibinfo{journal}{\emph{IEEE Transactions on Industrial Informatics}}
  (\bibinfo{year}{2022}).
\newblock


\bibitem[\protect\citeauthoryear{Xu, Zeng, Li, and Zheng}{Xu
  et~al\mbox{.}}{2022}]%
        {xu2022mfgan}
\bibfield{author}{\bibinfo{person}{Liming Xu}, \bibinfo{person}{Xianhua Zeng},
  \bibinfo{person}{Weisheng Li}, {and} \bibinfo{person}{Bochuan Zheng}.}
  \bibinfo{year}{2022}\natexlab{}.
\newblock \showarticletitle{MFGAN: Multi-modal Feature-fusion for CT Metal
  Artifact Reduction Using GANs}.
\newblock \bibinfo{journal}{\emph{ACM Transactions on Multimedia Computing,
  Communications, and Applications (TOMM)}} (\bibinfo{year}{2022}).
\newblock


\bibitem[\protect\citeauthoryear{Zhang, Li, Wang, Lu, Ma, and Qiu}{Zhang
  et~al\mbox{.}}{2020}]%
        {zhang2020psac}
\bibfield{author}{\bibinfo{person}{Yin Zhang}, \bibinfo{person}{Yujie Li},
  \bibinfo{person}{Ranran Wang}, \bibinfo{person}{Jianmin Lu},
  \bibinfo{person}{Xiao Ma}, {and} \bibinfo{person}{Meikang Qiu}.}
  \bibinfo{year}{2020}\natexlab{}.
\newblock \showarticletitle{PSAC: Proactive sequence-aware content caching via
  deep learning at the network edge}.
\newblock \bibinfo{journal}{\emph{IEEE Transactions on Network Science and
  Engineering}} \bibinfo{volume}{7}, \bibinfo{number}{4}
  (\bibinfo{year}{2020}), \bibinfo{pages}{2145--2154}.
\newblock


\bibitem[\protect\citeauthoryear{Zhen, Wang, Zhang, Yan, Wang, Ji, and
  Chen}{Zhen et~al\mbox{.}}{2022}]%
        {zhen2022towards}
\bibfield{author}{\bibinfo{person}{Peining Zhen}, \bibinfo{person}{Shuqi Wang},
  \bibinfo{person}{Suming Zhang}, \bibinfo{person}{Xiaotao Yan},
  \bibinfo{person}{Wei Wang}, \bibinfo{person}{Zhigang Ji}, {and}
  \bibinfo{person}{Hai-Bao Chen}.} \bibinfo{year}{2022}\natexlab{}.
\newblock \showarticletitle{Towards Accurate Oriented Object Detection in
  Aerial Images with Adaptive Multi-level Feature Fusion}.
\newblock \bibinfo{journal}{\emph{ACM Transactions on Multimedia Computing,
  Communications, and Applications (TOMM)}} (\bibinfo{year}{2022}).
\newblock


\bibitem[\protect\citeauthoryear{Zhou, Wu, Zhang, Kang, Ge, and Latecki}{Zhou
  et~al\mbox{.}}{2022a}]%
        {zhou2022contextual}
\bibfield{author}{\bibinfo{person}{Quan Zhou}, \bibinfo{person}{Xiaofu Wu},
  \bibinfo{person}{Suofei Zhang}, \bibinfo{person}{Bin Kang},
  \bibinfo{person}{Zongyuan Ge}, {and} \bibinfo{person}{Longin~Jan Latecki}.}
  \bibinfo{year}{2022}\natexlab{a}.
\newblock \showarticletitle{Contextual ensemble network for semantic
  segmentation}.
\newblock \bibinfo{journal}{\emph{Pattern Recognition}}  \bibinfo{volume}{122}
  (\bibinfo{year}{2022}), \bibinfo{pages}{108290}.
\newblock


\bibitem[\protect\citeauthoryear{Zhou, Yang, Gao, Ou, Lu, Chen, and
  Latecki}{Zhou et~al\mbox{.}}{2019}]%
        {zhou2019multi}
\bibfield{author}{\bibinfo{person}{Quan Zhou}, \bibinfo{person}{Wenbing Yang},
  \bibinfo{person}{Guangwei Gao}, \bibinfo{person}{Weihua Ou},
  \bibinfo{person}{Huimin Lu}, \bibinfo{person}{Jie Chen}, {and}
  \bibinfo{person}{Longin~Jan Latecki}.} \bibinfo{year}{2019}\natexlab{}.
\newblock \showarticletitle{Multi-scale deep context convolutional neural
  networks for semantic segmentation}.
\newblock \bibinfo{journal}{\emph{World Wide Web}} \bibinfo{volume}{22},
  \bibinfo{number}{2} (\bibinfo{year}{2019}), \bibinfo{pages}{555--570}.
\newblock


\bibitem[\protect\citeauthoryear{Zhou, Xia, Dou, Su, and Hu}{Zhou
  et~al\mbox{.}}{2022b}]%
        {zhou2022double}
\bibfield{author}{\bibinfo{person}{Wei Zhou}, \bibinfo{person}{Zhiwu Xia},
  \bibinfo{person}{Peng Dou}, \bibinfo{person}{Tao Su}, {and}
  \bibinfo{person}{Haifeng Hu}.} \bibinfo{year}{2022}\natexlab{b}.
\newblock \showarticletitle{Double Attention based on Graph Attention Network
  for Image Multi-Label Classification}.
\newblock \bibinfo{journal}{\emph{ACM Transactions on Multimedia Computing,
  Communications, and Applications (TOMM)}} (\bibinfo{year}{2022}).
\newblock


\end{thebibliography}

\end{document}